\documentclass[runningheads]{llncs}
\usepackage{graphicx}

\usepackage{tikz}
\usepackage{comment}
\usepackage{amsmath,amssymb} 
\usepackage{color}
\usepackage{bbding}

\usepackage[accsupp]{axessibility}  

\usepackage{cite}
\usepackage{mathrsfs}
\usepackage{xcolor}
\usepackage{comment}
\usepackage{pifont}

\usepackage{color}
\usepackage{hyperref}
\definecolor{myGreen}{RGB}{55,149,73}

\newcommand{\cmark}{\textcolor{green}{\ding{51}}}

\begin{document}
\pagestyle{headings}
\mainmatter
\def\ECCVSubNumber{00000}  

\title{You Should Look at All Objects} 

\titlerunning{YSLAO}
%
\author{
  Zhenchao Jin\inst{1} \and
  Dongdong Yu\inst{2} \and
  Luchuan Song\inst{3} \and
  Zehuan Yuan\inst{2} \and
  Lequan Yu\inst{1} \thanks{Corresponding author.}
}
\authorrunning{Zhenchao Jin, et al.}
%
\institute{
  The University of Hong Kong \\
  \email{blwx96@connect.hku.hk, lqyu@hku.hk} \and
  Bytedance \\
  \email{\{yudongdong, yuanzehuan\}@bytedance.com} \and
  University of Rochester \\
  \email{lsong11@ur.rochester.edu}
}

\maketitle

\begin{abstract}
  Feature pyramid network (FPN) is one of the key components for object detectors.
  However, there is a long-standing puzzle for researchers that the detection performance of large-scale objects are usually suppressed after introducing FPN.
  To this end, this paper first revisits FPN in the detection framework and reveals the nature of the success of FPN from the perspective of optimization.
  Then, we point out that the degraded performance of large-scale objects is due to the arising of improper back-propagation paths after integrating FPN. 
  It makes each level of the backbone network only has the ability to look at the objects within a certain scale range.  
  Based on these analysis, two feasible strategies are proposed to enable each level of the backbone to look at all objects in the FPN-based detection frameworks.
  Specifically, one is to introduce auxiliary objective functions to make each backbone level directly receive the back-propagation signals of various-scale objects during training.
  The other is to construct the feature pyramid in a more reasonable way to avoid the irrational back-propagation paths.
  Extensive experiments on the COCO benchmark validate the soundness of our analysis and the effectiveness of our methods. 
  Without bells and whistles, we demonstrate that our method achieves solid improvements (more than $2\%$) on various detection frameworks: 
  one-stage, two-stage, anchor-based, anchor-free and transformer-based detectors \footnote{Our code will be available at \href{https://github.com/CharlesPikachu/YSLAO}{https://github.com/CharlesPikachu/YSLAO}.}.
\end{abstract}

\section{Introduction}
\begin{figure*}[t]
\centering
\includegraphics[width=0.95\textwidth]{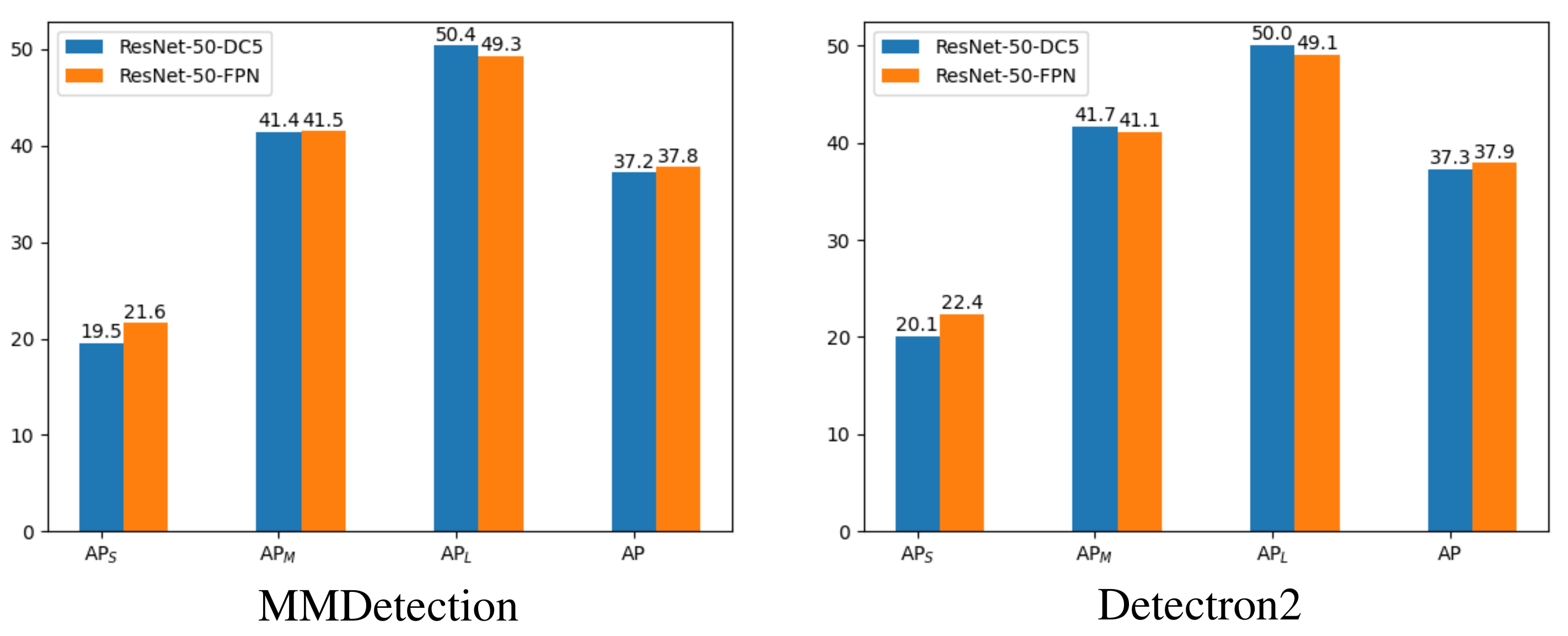}
\vspace{-0.40cm}
\caption{
  Comparing the detection performance between ResNet-50-DC5 and ResNet-50-FPN based on MMDetection \cite{chen2019mmdetection} and Detectron2 \cite{wu2019detectron2} toolboxes.
  The adopted detector is Faster R-CNN \cite{ren2015faster}. 
  The detectors are trained on COCO 2017 train set and evaluated on COCO 2017 validation set \cite{lin2014microsoft}.
}\label{fig.c5andfpn}
\vspace{-0.40cm}
\end{figure*}

Along with the advances in deep neural networks, recent years have seen remarkable progress in object detection, which aims at detecting objects of predefined categories.
A common belief for the success of the state-of-the-art detectors~\cite{he2017mask, jin2020safnet, zhang2020bridging, tian2019fcos, cai2019cascade} is the use of feature pyramid network (FPN) \cite{lin2017feature}.
Despite impressive, there is an unexpected phenomenon after introducing FPN that the overall detection performance improvement is built upon the \textit{increased} Average Precision of small objects (AP$_S$) and the \textit{decreased} Average Precision of large objects (AP$_L$).
For instance, the experiments based on MMDetection \cite{chen2019mmdetection} and Detectron2 \cite{wu2019detectron2} in Figure \ref{fig.c5andfpn} demonstrate this phenomenon.
When we leverage the detection toolbox MMDetection, we can observe that AP$_{S}$ increases from $19.5\%$ to $21.6\%$ while AP$_L$ decreases from $50.4\%$ to $49.3\%$ after integrating FPN.
The consistent tend can also be observed in Detectron2.

Prior to this study, there are mainly two assumptions on why the introduction of FPN works.
The first is that the use of FPN helps obtain better representations by fusing multiple low-level and high-level feature maps \cite{lin2017feature, liu2018path, kong2018deep, pang2019libra, ghiasi2019fpn}.
The second is that each pyramid level can be responsible for detecting objects within a certain scale range, \emph{i.e.}, divide-and-conquer \cite{chen2021you}.
Obviously, both assumptions should lead to the same conclusion that the increase in AP is due to the co-increase in AP$_S$, AP$_M$ and AP$_L$.
However, the unexpected drops of AP$_L$ in Figure \ref{fig.c5andfpn} indicates that there are other key differences between FPN-free and FPN-based detection frameworks, while few studies have taken note of this.
In this paper, we propose to investigate FPN from the perspective of optimization.
Our assumption is that \emph{except for the multi-scale feature fusion and divide-and-conquer, the back-propagation paths altered by FPN will also directly influence the performance of the detection frameworks}.

We start from explaining why FPN can benefit the detection framework by changing the back-propagation paths.
Then, we point out that the back-propagation paths altered by the existing FPN paradigm will make each backbone stage only have the ability to see the objects within a certain scale range (\emph{i.e.}, extracting features that are only fit to certain scale range objects), which is the cause of the inconsistent changes in AP$_S$, AP$_M$ and AP$_L$ in Figure \ref{fig.c5andfpn}.
Accordingly, the key insight to achieve the consistent improvements in AP of the objects with various scale ranges is to enable each backbone stage to see all objects during training.
Based on this principle, we propose to expand and amend the existing back-propagation paths in FPN-based detection frameworks.

Our approach of expanding the back-propagation paths is to introduce auxiliary objective functions so that both the original signals and the extra signals can jointly supervise the learning of the corresponding backbone levels.
The key technique to the success of this approach is to introduce the uncertainty \cite{kendall2018multi, kendall2017uncertainties} to better balance the various back-propagation signals.
The strategy of amending the back-propagation paths is to build the feature pyramid in a more effective way and thereby, all levels of the backbone network can receive the sufficient signals.
The key technique of this approach is the feature grouping module used to promise the space compactness of homogeneous representations.

In a nutshell, the contributions of this paper are:

\begin{itemize}

  \item To the best of our knowledge, it is the first work to reveal the nature of the success of FPN from the perspective of optimization.
  Further, we provide new insight in explaining why the introduction of the traditional FPN would suppress the performance of large-scale objects from this perspective.

  \item We propose to introduce auxiliary objective functions guided by uncertainty to mitigate the inconsistent changes in AP$_S$, AP$_M$ and AP$_L$.
  Since there are no additional computational overhead during testing in the strategy, the inference speed of the detectors can be preserved from decreasing.
  
  \item We propose a novel feature pyramid generalization paradigm.
  The key idea is to make the back-propagation signals of various-scale objects can directly pass to each level of the backbone network.
  We further design a cascade structure to achieve more robust Average Precision (AP) improvements.

  \item The extensive experiments on COCO benchmark validate the soundness of our principle and the effectiveness of our solutions.
  Without bells and whistles, our method boosts the detection performance by more than $2\%$ AP on various frameworks: 
  one-stage, two-stage, anchor-based, anchor-free and transformer-based detectors.
  
\end{itemize}

\section{Related Works}

\noindent \textbf{Object Detection.}
Recent years have witnessed remarkable improvements in object detection \cite{cai2019cascade, tian2019fcos, jin2020safnet, pang2019libra, sun2021sparse}.
In general, there are two leading paradigms in this area, \emph{i.e.}, one-stage and two-stage frameworks.
Two-stage pipeline is first introduced by R-CNN \cite{girshick2014rich}, where a set of region proposals are yielded in the first stage, and then the second stage classifies and refines the proposals.
The next milestone of two-stage detector is the emergence of Faster R-CNN \cite{ren2015faster}, which aims to improve the efficiency of two-stage methods and allow the detectors to be trained end-to-end.
After that, plenty of algorithms have been proposed to further boost its performance, 
including applying multi-scale training and testing \cite{singh2018sniper, singh2018analysis}, 
redesigning and reforming architecture \cite{he2017mask, cai2019cascade, wu2020rethinking, zhu2019deformable, chen2019hybrid},
introducing relation and attention mechanism \cite{liu2018structure, hu2018relation, shrivastava2016contextual},
improving the training strategy and loss function \cite{shrivastava2016training, li2020generalized, he2019bounding, micikevicius2017mixed, qian2020dr}, 
adopting more reasonable post-processing algorithms \cite{bodla2017soft, wang2021end, he2018softer, liu2019adaptive}.
Different from the two-stage approaches, one-stage detectors directly predict the object category and location based on the predefined anchors.
They are simpler and faster than two-stage methods but have trailed the detection performance until the emergence of RetinaNet \cite{lin2017focal}.
Thereafter, lots of works \cite{chen2021you, ge2021yolox, tian2019fcos, li2019dynamic} are presented to boost the detection performance of one-stage detectors and at present, one-stage methods can achieve very close performance with two-stage frameworks at a faster inference speed.

\noindent \textbf{Feature Pyramids.}
Feature pyramids have dominated modern detectors for serval years.
Recent researches on feature pyramids can be roughly categorized into three gatherings:
top-down or bottom-up networks \cite{lin2017feature, liu2018path, shrivastava2016beyond, pang2019libra}, 
attention based methods \cite{jin2020safnet, zhang2020feature, zhao2021graphfpn, kong2018deep},
and neural architecture search based approaches \cite{ghiasi2019fpn, tan2020efficientdet}.
Specifically, feature pyramid network (FPN) \cite{lin2017feature} is one of the most classical paradigms to build a feature pyramid,
which designs a top-down architecture with lateral connections to make each pyramid level carry the high-level semantic information.
After that, several works \cite{liu2018path, kong2018deep, ghiasi2019fpn, pang2019libra, zhao2021graphfpn, jin2020safnet} follow FPN and make attempts to obtain more effective representations by improving the strategy of multi-scale feature fusion.
PANet \cite{liu2018path} proposes to leverage bottom-up architecture to shorten the information interaction path between shallow layers and topmost features.
SAFNet \cite{jin2020safnet} aims to suppress the redundant information at all pyramid scales by introducing attention mechanism.
Nas-fpn \cite{ghiasi2019fpn} proposes to construct the feature pyramids by neural architecture search.
However, the starting point of the above methods is that FPN can brings two benefits, \emph{i.e.}, leveraging multi-scale feature fusion to obtain more effective representations \cite{lin2017feature, liu2018path, kong2018deep, pang2019libra, ghiasi2019fpn} and adopting divide-and-conquer to reduce the learning difficulty \cite{chen2021you}.
And it fails to explain why introducing FPN will suppress the performance of large-scale objects.
Motivated by this, we propose to revisit FPN from the perspective of optimization, which successfully explains the anomalous phenomenon in Figure \ref{fig.c5andfpn}.
From this novel starting point, we further propose to mitigate the inconsistent changes in AP$_S$, AP$_M$ and AP$_L$ by expanding or amending the back-propagation paths in FPN-based detection frameworks.
And it is the main difference between our approaches and previous works.

\section{Revisit FPN}

\subsection{Backbone Network} \label{sec.backbone}
In object detection, the backbone network $\mathcal{F}_{B}$ is used to extract the basic features $\mathcal{C}$ from the input image $I$.
For the convenience of presentation, we assume that the adopted backbone network is ResNet \cite{he2016deep}.
It generally consists of one basic feature extractor and plenty of residual blocks, where the residual blocks can be grouped into four stages according to the resolutions of the output feature maps.
Specifically, $\mathcal{C}$ is calculated as follow,
\begin{equation} \label{eq.backbone}
\begin{aligned}
  C_1 &= f_{s_0}(I), \\
  C_i &= f_{s_{i-1}}(C_{i-1}),~ 2 \leq i \leq 5,
\end{aligned}
\end{equation}
where $\mathcal{C}$ consists of $\{C_2, C_3, C_4, C_5\}$ and $\mathcal{F}_{B}$ consists of $\{f_{s_0}, f_{s_1}, f_{s_2}, f_{s_3}, f_{s_4}\}$.

\subsection{FPN-free Detection Framework} \label{sec.fpnfree}
For FPN-free detectors, the network usually leverages $C_5$ to perform the classification and regression of the objects as follow,
\begin{equation} \label{eq.c5predict}
\begin{aligned}
  O_{cls} &= f_{cls}(f_{pre}(C_5)), \\
  O_{reg} &= f_{reg}(f_{pre}(C_5)),
\end{aligned}
\end{equation}
where $f_{pre}$ is introduced to unify various operations between $C_5$ and the output results, \emph{e.g.}, the region proposal network \cite{ren2015faster}.
$O_{cls}$ and $O_{reg}$ are the predicted category information and location information of the objects, respectively.
$f_{cls}$ and $f_{reg}$ are a $1 \times 1$ convolution layer, respectively.
During training, the classification and regression loss are calculated as follow,
\begin{equation} \label{eq.c5loss}
  L = L_{cls} (O_{cls}, GT_{cls}) + \lambda L_{reg} (O_{reg}, GT_{reg}),
\end{equation}
where the adopted objective functions $L_{cls}$ and $L_{reg}$ depend on the utilized detection framework.
$GT_{cls}$ and $GT_{reg}$ are the ground-truth classification and regression information, respectively.
$\lambda$ is a hyper-parameter used to balance the classification and regression losses.

\subsection{FPN-based Detection Framework} \label{sec.fpnbased}
For FPN-based detectors, $\mathcal{C}$ is first used to build the feature pyramid as follow,
\begin{equation} \label{eq.fpn}
\begin{aligned}
  C'_{5} &= f_{lat_5}(C_{5}), \\
  C'_{4} &= f_{lat_4}(C_{4}) + UP_{2 \times} (C'_{5}), \\
  C'_{3} &= f_{lat_3}(C_{3}) + UP_{2 \times} (C'_{4}), \\
  C'_{2} &= f_{lat_2}(C_{2}) + UP_{2 \times} (C'_{3}), \\
  P_{l} &= f_{smo_l} (C'_{l}),~ 2 \leq l \leq 5,
\end{aligned}
\end{equation}
where $\mathcal{P} = \{P_2, P_3, P_4, P_5\}$ is the constructed feature pyramid.
$UP_{2 \times}$ denotes for the upsampling with the scale factor of $2$.
$f_{lat_{i}}, 2 \leq i \leq 5$ is the lateral connections implemented by a $1 \times 1$ convolution layer, respectively, 
which is used to change the number of the channels of $\mathcal{C}$.
$f_{smo_{l}}, 2 \leq l \leq 5$ is a linear function and is usually implemented by a $3 \times 3$ convolution layer.
Without loss of generality, Eq.(\ref{eq.fpn}) can be rewritten as follow,
\begin{equation} \label{eq.rewrittenfpn}
  P_{l} = \sum_{i=l}^{5} w_{i} \cdot C_{i}, ~ 2 \leq l \leq 5,
\end{equation}
where $w_{i}$ is the final weights for the correspond level after polynomial expansions \cite{kong2018deep}.
Then, the network uses $\mathcal{P}$ to predict the classification and regression information of the objects assigned to each pyramid level $l$ as follow, 
\begin{equation} \label{eq.fpnpredict}
\begin{aligned}
  O_{cls, l} &= f_{cls, l}(f_{pre, l}(P_l)), \\
  O_{reg, l} &= f_{reg, l}(f_{pre, l}(P_l)).
\end{aligned}
\end{equation}
The objects assignment rule is to make the low-resolution pyramid features (\emph{e.g.}, $P_5$) be responsible for predicting the large-scale objects,
while the high-resolution pyramid features (\emph{e.g.}, $P_2$) are utilized to predict the small-scale objects.
During network optimization, the losses at each pyramid level $l$ are calculated as follow,
\begin{equation} \label{eq.fpnloss}
  L_l = L_{cls} (O_{cls, l}, GT_{cls, l}) + \lambda L_{reg} (O_{reg, l}, GT_{reg, l}).
\end{equation}

\begin{figure*}[t]
\centering
\includegraphics[width=0.7\textwidth]{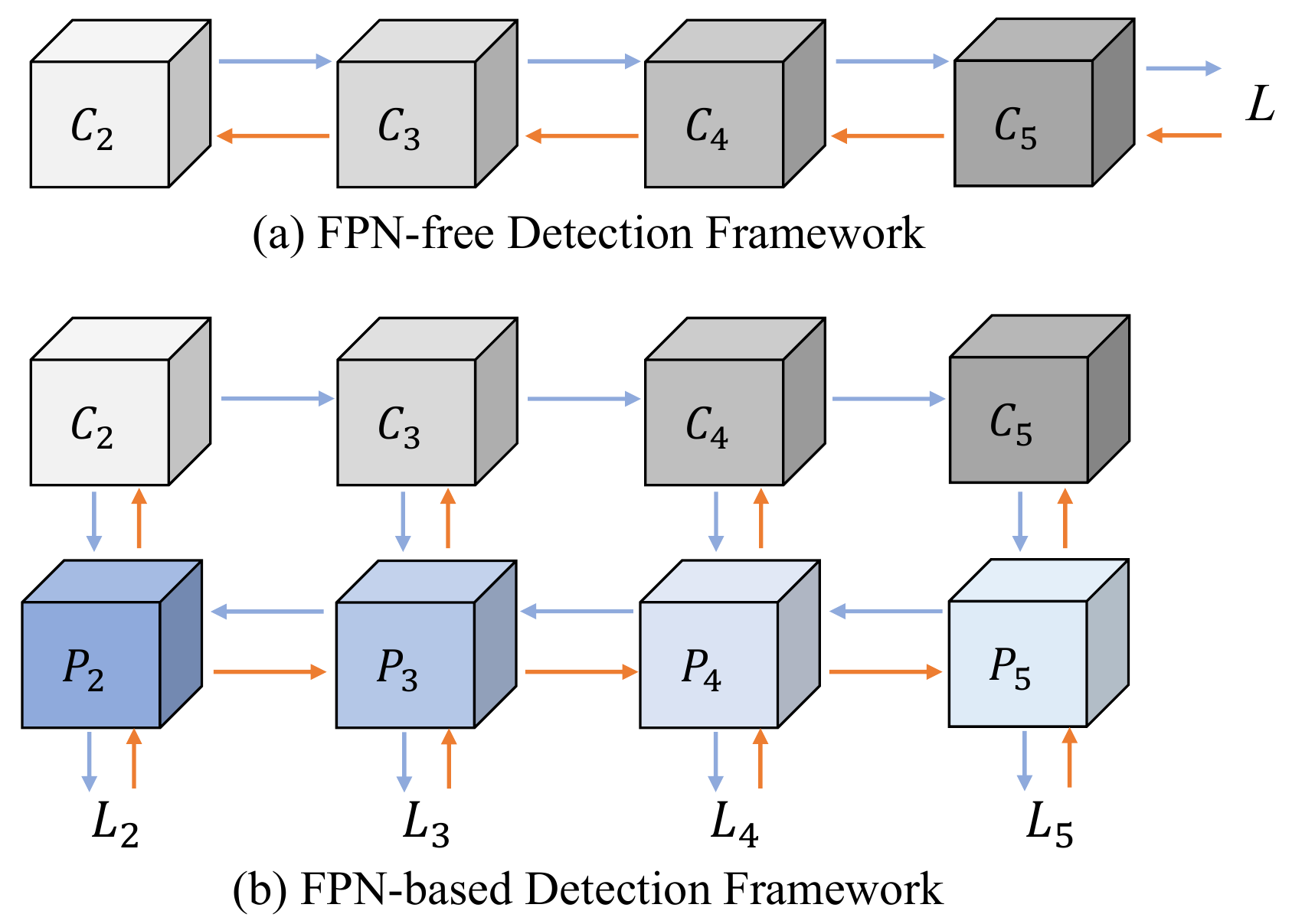}
\vspace{-0.40cm}
\caption{
  Comparing the back-propagation paths between FPN-free detection framework and FPN-based detection framework.
  The blue arrows represent for forward and the orange arrows denote for back propagation.
  Note that, only the most significant back-propagation signals to each backbone level will be marked.
}\label{fig.lossforc5andfpn}
\vspace{-0.40cm}
\end{figure*}

\subsection{Analysis of FPN} \label{sec.analysisfpn}
From Section \ref{sec.fpnfree} and \ref{sec.fpnbased}, we can observe that introducing FPN can alter the back-propagation paths between the objective functions and the backbone network.
Figure \ref{fig.lossforc5andfpn} shows the detailed differences between the FPN-free and FPN-based detection framework.
In the FPN-free detection pipeline, only the backbone feature $C_5$ is directly under the supervision of the objective functions.
Since there exists the vanishing gradient problem in deep neural networks, the shallow layers (\emph{i.e.}, $\{f_{s_0}, f_{s_1}, f_{s_2}, f_{s_3}\}$) of the backbone network 
will be difficult to receive effective supervision by the backward propagation. 
While in the FPN-based detection framework, we can observe that all the backbone features are directly under the supervision of the objective functions. 
Since this strategy avoids the vanishing gradient problem for the shallow layers, each level of the backbone network can receive more supervision to train its own parameters.
We believe that it is the key reason why FPN-based detectors outperform FPN-free detectors from the perspective of optimization.

\begin{figure*}[t]
\centering
\includegraphics[width=0.7\textwidth]{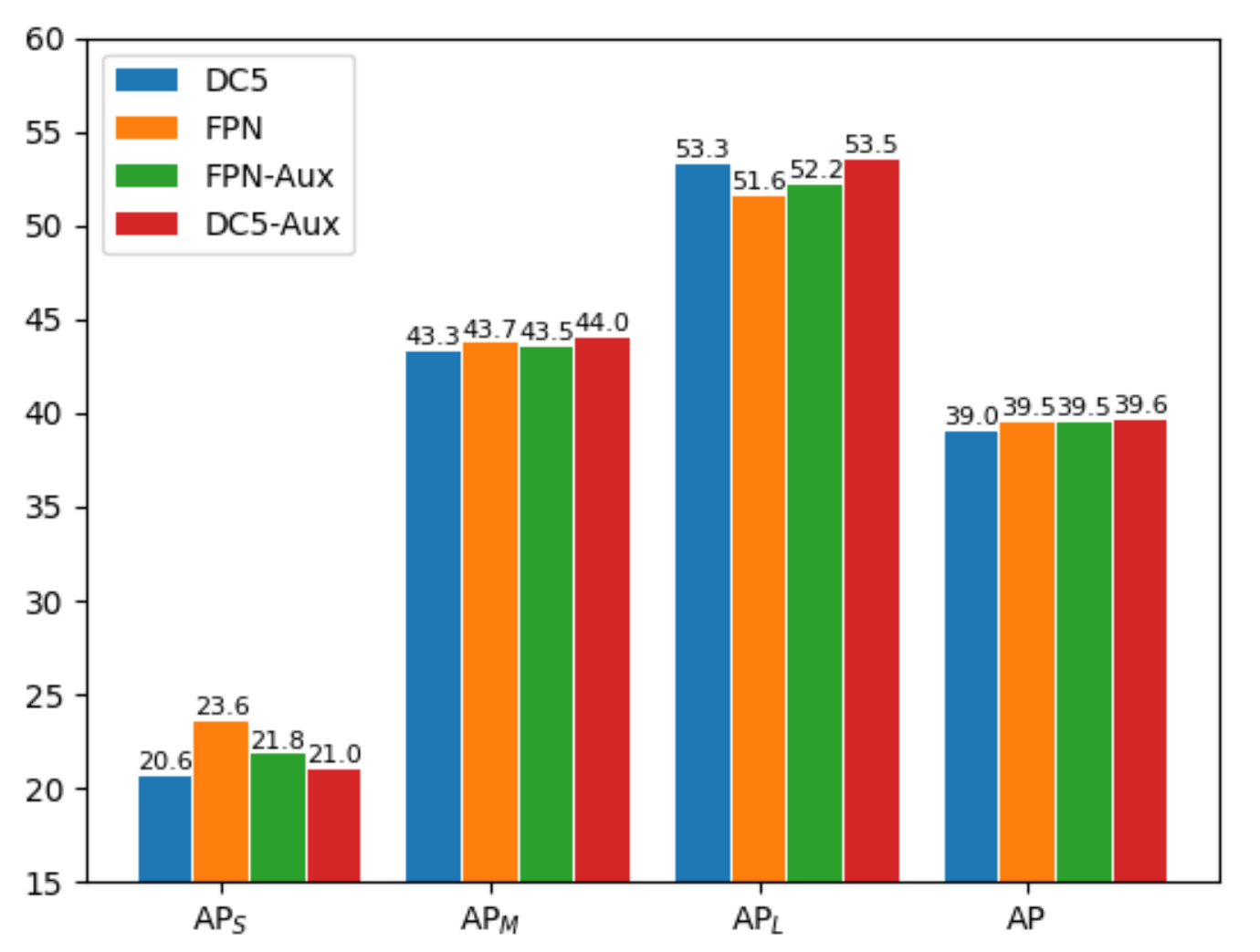}
\vspace{-0.40cm}
\caption{
  The detection performance with different settings.
  The adopted backbone network is ResNet-101 and the utilized detector is Faster R-CNN.
  The models are trained on COCO 2017 train set and evaluated on COCO 2017 validation set \cite{lin2014microsoft}.
}\label{fig.preliminaryfortheory}
\vspace{-0.40cm}
\end{figure*}

To further demonstrate the principle above, we conduct the empirical study and show the experimental results in Figure \ref{fig.preliminaryfortheory}.
FPN-Aux and DC5-Aux denote for introducing the auxiliary losses in the shallow layers of the backbone networks~\cite{zhao2017pyramid, wang2021end}.
Specifically, given $\mathcal{C}$, we first have
\begin{equation}
\begin{aligned}
  \hat{O}_{cls, i} &= \hat{f}_{cls, i}(\hat{f}_{pre, i}(C_i)), ~ 2 \leq i \leq 4, \\
  \hat{O}_{reg, i} &= \hat{f}_{reg, i}(\hat{f}_{pre, i}(C_i)), ~ 2 \leq i \leq 4.
\end{aligned}
\end{equation}
For the two-stage detectors \cite{ren2015faster}, to avoid double calculation of the proposals, we will utilize the proposals calculated in Eq.(\ref{eq.c5predict}) or Eq.(\ref{eq.fpnpredict}) to extract the ROIs.
Then, the auxiliary losses can be obtained as follow,
\begin{equation}
  \hat{L}_{i} = L_{cls}(\hat{O}_{cls, i}, GT_{cls}) + \lambda L_{reg}(\hat{O}_{reg, i}, GT_{reg}).
\end{equation}
And the final loss of the detection framework is the summation of the auxiliary losses and the original losses.
Since the auxiliary losses can be used to directly supervise the learning of the shallow layers of the backbone network, if our assumption is correct, introducing auxiliary losses should own a similar function to integrating FPN from the perspective of optimization.
In Figure \ref{fig.preliminaryfortheory}, it is observed that the auxiliary losses can boost the detection performance of FPN-free detector (from $39.0\%$ to $39.6\%$) and obtain a comparable AP result to FPN-based detector ($39.6\%$ \emph{v.s.} $39.5\%$).
However, the introduction of auxiliary losses seems useless to FPN-based detector (from $39.5\%$ to $39.5\%$). 
This result validates our assumption that from the perspective of optimization, the nature of the success of FPN is the shorten back-propagation distance between the objective losses and the shallow layers of the backbone network.

Now, the question is why the introduction of FPN will suppress the detection performance of large-scale objects.
As illustrated in Figure \ref{fig.lossforc5andfpn}, $P_2$ is a linear combination of $\{C_2, C_3, C_4, C_5\}$, therefore, $L_2$ can directly supervise the learning of all the backbone stages.
With the similar principle, $L_3$, $L_4$ and $L_5$ have the ability to directly constrain $\{C_3, C_4, C_5\}$, $\{C_4, C_5\}$ and $C_5$, respectively.
However, as mentioned above, $L_{2}$ is only used to make the corresponding backbone levels focus on the objects within a small scale range.
Thus, the learned feature $C_2$ only has the ability to detect the small-scale objects well by back propagation.
Meanwhile, the backbone network also needs to utilize Eq.(\ref{eq.backbone}) to calculate $\{C_3, C_4, C_5\}$ with $C_{2}$ as input.
Obviously, it is insufficient for $f_{s_2}$ to extract rich semantic features of the larger objects from $C_2$.
Worse still, the adverse effects will be further accumulated when leveraging $f_{s_3}$ and $f_{s_4}$ to calculate $C_4$ and $C_5$.
As a result, the semantic information carried by $C_5$ is somehow ineffective for predicting the large-scale objects.
And it is why there is always an unexpected phenomenon after introducing FPN that the overall detection performance improvement is built upon the increased AP$_S$ and the decreased AP$_L$.

The empirical study in Figure \ref{fig.preliminaryfortheory} also validate our assumption.
In detail, we can observe that after applying the auxiliary losses to the FPN-based detection frameworks, the performance improvements among $\{\text{AP}_S,\text{AP}_M,\text{AP}_L\}$ tend to be consistent with FPN-free detectors w/ the auxiliary losses.
The result shows that as the auxiliary losses can help the shallow layers of the backbone learn the features to detect various-scale objects, $C_5$ no longer suffers from the ineffective features for predicting large-scale objects as only integrating FPN into the detection framework.
In other words, the lack of effective semantic information of large-scale objects in $C_5$ is the key reason for the decrease of AP$_L$.
And the problem is derived from the inability of $f_{s_{i}},1 \leq i \leq 3$ to look at various-scale objects during training.

\section{Methodology}
Motivated by the finding that the inconsistent changes in $\{\text{AP}_S,\text{AP}_M,\text{AP}_L\}$ is caused by the inability of $f_{s_{i}},1 \leq i \leq 3$ to see all objects during training,
we propose to make the backbone stages look at various-scale objects by expanding or amending the back-propagation paths in FPN-based detection frameworks to address the decreased AP$_L$ problem above.
Specifically, we propose two strategies, \emph{i.e.}, introducing auxiliary objective functions and building the feature pyramid in a more reasonable manner in this section.

\subsection{Auxiliary Losses}
As mentioned in Section \ref{sec.analysisfpn}, introducing auxiliary losses can help $f_{s_{i}},1 \leq i \leq 3$ own the ability to see all objects.
However, the simple summation of the losses may be insufficient.
In order to introduce auxiliary losses more rationally, we propose to leverage the uncertainty \cite{kendall2018multi, kendall2017uncertainties, yang2021uncertainty} to better balance the various-type loss signals.
Specifically, we incorporate the uncertainty into each classification and regression auxiliary loss as follow,
\begin{equation}
  \mathcal{L} (p, gt) = e^{-\alpha} \hat{\mathcal{L}}(p, gt) + \tau \alpha,
\end{equation}
where $p$ is the predicted result and $gt$ is the corresponding ground truth.
$\hat{\mathcal{L}}$ denotes for the loss function, \emph{e.g.}, $L_{reg}$ and $L_{cls}$.
$\tau$ is a hyper-parameter used to avoid generating high uncertainty $\alpha$.
$\alpha$ is generated as follow,
\begin{equation}
  \alpha = ReLU(w \cdot x + b),
\end{equation}
where $x$ is the feature map also utilized to predict $p$. 
$w$ and $b$ are the learnable parameters.
$ReLU$ is used to promise $\alpha \geq 0$.

\subsection{Feature Pyramid Generation Paradigm}

Building the feature pyramid in a more reasonable way is also an effective method to achieve consistent improvements in $\{\text{AP}_S,\text{AP}_M,\text{AP}_L\}$.
As analysed in Section \ref{sec.analysisfpn}, the problem in the process of constructing traditional FPN is caused by Eq.(\ref{eq.fpn}).
Specifically, $P_l$ should contain the feature maps from all backbone levels so that $L_l$ can help each backbone level see the objects inputted to $L_l$.
Accordingly, the summation of $L_l, 2 \leq l \leq 5$ can make each backbone level own the ability to look at all objects.

\begin{figure}
\centering
\includegraphics[width=0.38\textwidth]{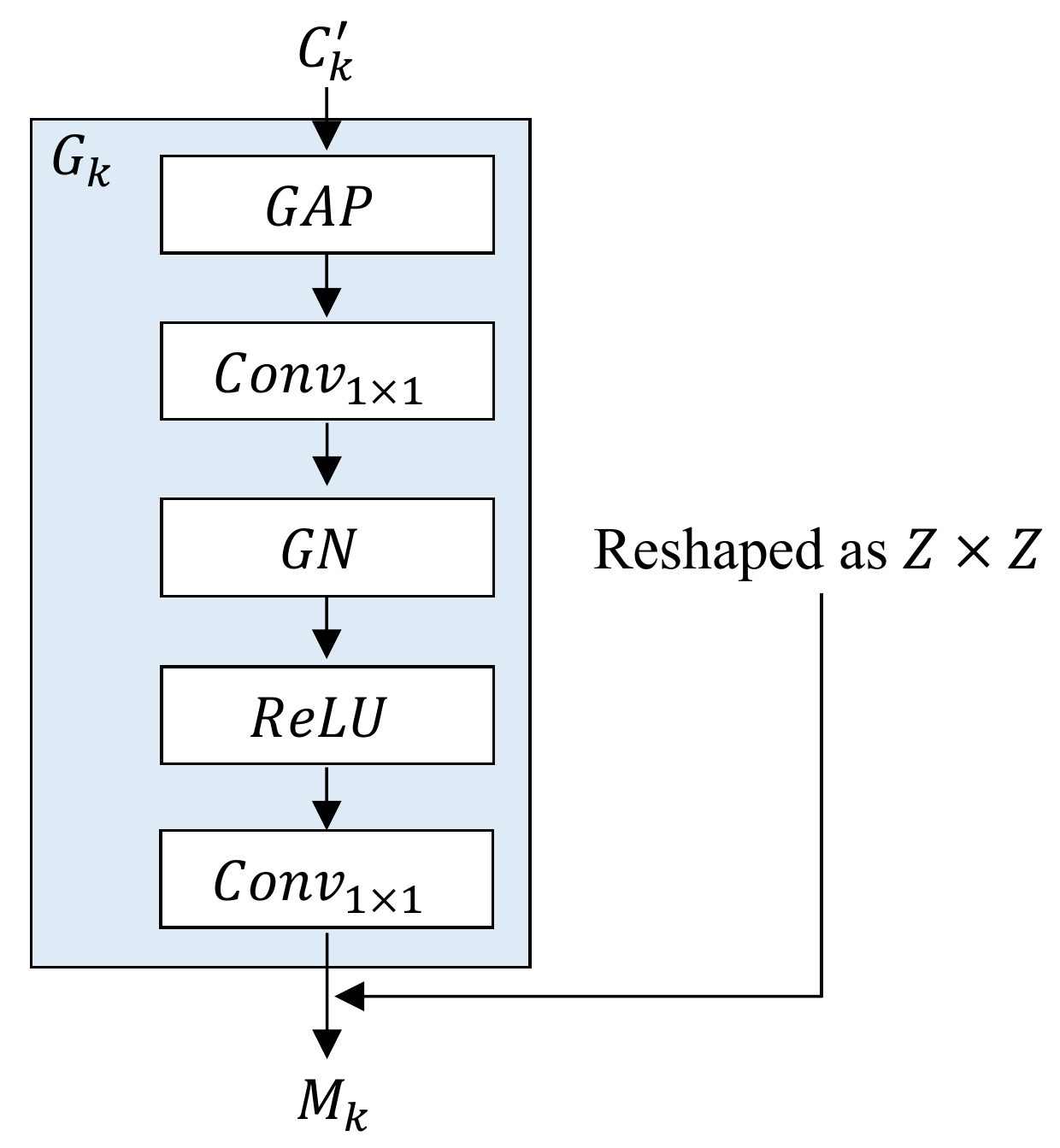}
\vspace{-0.45cm}
\caption{
  An illustration of the structure of $G_k$.
  $Conv_{1 \times 1}$ denotes for a $1 \times 1$ convolution layer.
  $GN$ means group normalization and $GAP$ is the global average pooling operation.
}\label{fig.Gkstructure}
\vspace{-0.45cm}
\end{figure}

\noindent \textbf{Feature Grouping.} 
To select effective feature maps from $\mathcal{C}'=\{C'_2, C'_3, C'_4, C'_5\}$ for the objects assigned to the corresponding pyramid level, we first perform channel swapping on $\mathcal{C}'$ as follow,
\begin{equation}
  X_k = R^{zhw}(M_k \otimes R^{zn}(C'_k)),~2 \leq k \leq 5,
\end{equation}
where $\otimes$ denotes for matrix multiplication.
$R^{zn}$ reshapes $C'_k$ into the size of $Z \times HW$ and $R^{zhw}$ reshapes the input tensor into the size of $Z \times H \times W$,
where $Z$ is the number of the channels and $H \times W$ is the resolution of the feature map.
$M_k$ is a matrix of size $Z \times Z$ used to achieve channel swapping.
In practice, $M_k$ is generated as follow,
\begin{equation}
  M_{k} = G_k (C'_k),
\end{equation}
where the structure of $G_k$ is shown in Figure \ref{fig.Gkstructure}.
We expect $M_{k}$ to own the ability to make the homogeneous feature maps become compact along the channel dimension.

Then, $X_k$ is divided into quarters along the channel dimension,
\begin{equation}
  X_{k} = \{X_{k,2}, X_{k,3}, X_{k,4}, X_{k,5}\},
\end{equation}
where we assume that $X_{k, l}, 2 \leq l \leq 5$ only carries the effective semantic information of the objects assigned to the pyramid level $l$.
After that, we have
\begin{equation}
  P'_{l} = X_{2,l} \oplus X_{3,l} \oplus X_{4,l} \oplus X_{5,l},~ 2 \leq l \leq 5,
\end{equation}
where $\oplus$ denotes for the concatenation operation.
Finally, the feature pyramid is constructed as follow,
\begin{equation} \label{eq.fgfpn}
  P_{l} = f_{smo_l} (P'_{l}),~ 2 \leq l \leq 5.
\end{equation}

\noindent \textbf{Cascade Structure.} 
To better promote the space compactness, we propose to employ a cascade structure to conduct feature grouping in a coarse-to-fine manner.
Specifically, at the second stage, $P'_{l}$ will first be taken as the input of the feature grouping module and thereby we can obtain $\hat{P}'_{l}$.
Then, we have
\begin{equation}
  P''_{l} = f_w(P'_{l}) \cdot P'_{l} + f_w(\hat{P}'_{l}) \cdot \hat{P}'_{l},~ 2 \leq l \leq 5,
\end{equation}
where $f_w$ is a non-linear function used to generate the feature fusion weights. In our implementation, $f_w$ consists of two convolution blocks (a block consists of a convolution, a normalization and an activation layer).
Finally, Eq.(\ref{eq.fgfpn}) will be conducted to obtain the feature pyramid with the input $P''_{l}$.
The same can be done for the cases when the number of the stages is greater than $2$.

\section{Experiments}

\noindent \textbf{Dataset.} 
Our approaches are evaluated on the challenging MS COCO benchmark~\cite{lin2014microsoft}, which contains $\sim$118k images for training (\emph{train-2017}), 5k images for validation (\emph{val-2017}) and $\sim$20k images with no disclosed annotations for testing (\emph{test-dev}).
By default, the detection frameworks in this section are trained on \emph{train-2017} set and evaluated on \emph{val-2017} set. 

\noindent \textbf{Implementation Details.}
Our methods are implemented with MMDetection \cite{chen2019mmdetection}.
We train our detection frameworks on 8 NVIDIA Tesla V100 GPUs with a 32 GB memory per-card.
Following previous works~\cite{lin2017feature, ren2015faster, lin2017focal}, we initialize the backbone networks using the weights pre-trained on ImageNet \cite{krizhevsky2012imagenet} and randomly initialize the weights of the newly added modules.
The input images are resized to keep their shorter side being 800 and their longer side less or equal to 1,333.
The optimizer is stochastic gradient descent (SGD) with momentum of 0.9, weight decay of 0.0001, and batch size of 16 (\emph{i.e.}, 2 images per GPU).
By default, the models are trained for 12 epochs ($1\times$ schedule), and we set the initial learning rate as 0.02 and decay it by 0.1 at epoch 9 and 11, respectively.
We adopt random horizontal flip as the data augmentation.
Other unmentioned hyper-parameters follow the settings in MMDetection.

In the inference phase, the input image is first resized in the same way as the training phase and then we forward it through the whole network to output the predicted bounding boxes with the category probability distribution.
After that, we leverage a score 0.05 to preliminary filter out background bounding boxes and then output the top 1,000 detections per pyramid level.
Finally, the non-maximum suppression (NMS) is applied with the IoU threshold 0.5 per class to output the final top 100 confident detections per image.

\noindent \textbf{Evaluation Metrics.}
The results are evaluated with standard COCO-style metrics, including AP (averaged over IoU thresholds), AP$_{50}$ (AP for IoU threshold $50\%$), AP$_{75}$ (AP for IoU threshold $75\%$), AP$_{S}$ (AP on objects of small scales), AP$_{M}$ (AP on objects of medium scales) and AP$_{L}$ (AP on objects of large scales).

\begin{table*}[t]
\centering
\caption{
  Ablation study on auxiliary losses. FPS is evaluated on a single Titan Xp.
}\label{table.ablationaux}
\vspace{-0.2cm}
\resizebox{.95\textwidth}{!}{
\begin{tabular}{c|c|c|c|ccc|ccc|cccc}
  \hline
  \hline
  Framework                              &Backbone     &Auxiliary      &Uncertainty    &AP         &AP$_{50}$       &AP$_{75}$      &AP$_{S}$      &AP$_{M}$     &AP$_{L}$    &FPS  \\
  \hline
  \emph{One-stage} &&&&&&&&&&&\\
  RetinaNet                              &ResNet-101   &               &               &38.5       &57.6            &41.0           &21.7          &42.8         &50.4        &15.0 \\
  RetinaNet                              &ResNet-101   &\cmark         &               &38.7       &57.7            &41.2           &21.2          &43.0         &51.2        &15.0 \\
  RetinaNet                              &ResNet-101   &\cmark         &\cmark         &40.1       &61.4            &43.7           &23.3          &44.4         &52.4        &15.0 \\
  \hline
  \emph{Two-stage} &&&&&&&&&&&\\
  Faster R-CNN                           &ResNet-101   &               &               &39.5       &60.4            &42.9           &23.6          &43.7         &51.6        &15.6 \\
  Faster R-CNN                           &ResNet-101   &\cmark         &               &39.5       &60.0            &43.3           &21.8          &43.5         &52.2        &15.6 \\
  Faster R-CNN                           &ResNet-101   &\cmark         &\cmark         &40.9       &62.0            &44.8           &24.2          &45.3         &53.3        &15.6 \\
  \hline
  \hline
\end{tabular}}
\end{table*}

\begin{table*}[t]
\centering
\caption{
  Ablation study on the feature pyramid generation paradigm. 
}\label{table.ablationfg}
\vspace{-0.2cm}
\resizebox{.95\textwidth}{!}{
\begin{tabular}{c|c|c|c|ccc|ccc|cccc}
  \hline
  \hline
  Framework                              &Backbone     &Feature Grouping   &Cascade Times    &AP         &AP$_{50}$       &AP$_{75}$      &AP$_{S}$      &AP$_{M}$     &AP$_{L}$   &FPS  \\
  \hline
  \emph{One-stage} &&&&&&&&&&&\\
  RetinaNet                              &ResNet-101   &                   &                 &38.5       &57.6            &41.0           &21.7          &42.8         &50.4       &15.0 \\
  RetinaNet                              &ResNet-101   &\cmark             &$1 \times$       &40.2       &60.1            &42.6           &23.3          &44.5         &52.7       &12.2 \\
  RetinaNet                              &ResNet-101   &\cmark             &$2 \times$       &40.8       &60.5            &43.6           &24.0          &44.8         &54.4       &11.7 \\
  RetinaNet                              &ResNet-101   &\cmark             &$3 \times$       &41.2       &60.8            &43.8           &24.1          &45.2         &55.2       &11.1 \\
  \hline
  \emph{Two-stage} &&&&&&&&&&&\\
  Faster R-CNN                           &ResNet-101   &                   &                 &39.5       &60.4            &42.9           &23.6          &43.7         &51.6       &15.6  \\
  Faster R-CNN                           &ResNet-101   &\cmark             &$1 \times$       &40.6       &61.9            &44.5           &24.2          &45.0         &52.7       &12.0  \\
  Faster R-CNN                           &ResNet-101   &\cmark             &$2 \times$       &41.7       &62.7            &45.4           &24.8          &45.9         &53.7       &11.4  \\
  Faster R-CNN                           &ResNet-101   &\cmark             &$3 \times$       &42.2       &63.0            &45.8           &25.5          &46.1         &55.8       &10.9  \\
  \hline
  \hline
\end{tabular}}
\vspace{-0.2cm}
\end{table*}

\subsection{Ablation Studies}

\noindent \textbf{Auxiliary Losses.}
Since the auxiliary losses can build extra back-propagation paths between the objective functions and the backbone levels, 
we propose to introduce auxiliary losses to address the dropped AP$_L$ problem.
Table \ref{table.ablationaux} shows the ablation experiments.
After introducing the auxiliary losses, it is observed that AP$_L$ increases from $50.4\%$ to $51.2\%$ and from $51.6\%$ to $52.2\%$ in the one-stage and two-stage detector, respectively.
The improvements indicate that the auxiliary losses can help $f_{s_{i}},1 \leq i \leq 3$ own the ability to see all objects and thereby $C_5$ can carry more effective semantic information of the large-scale objects.
Furthermore, we can observe that AP$_S$ drops a lot if simply add the auxiliary losses and the original losses linearly.
The drops indicate that to some extent, the auxiliary signals will overwrite the original loss signals especially for the small objects whose effective information is the least.
To this end, we introduce the uncertainty to the auxiliary losses to scale the auxiliary signals adaptively.
It is observed that AP$_S$ improves from $21.2\%$ to $23.3\%$ and from $21.8\%$ to $24.2\%$ in the one-stage and two-stage pipeline, respectively.
As a result, the overall detection performance improves with the consistent changes in $\{\text{AP}_S,\text{AP}_M,\text{AP}_L\}$.
These improvements well demonstrate the correctness of our speculation and the effectiveness of our method.
Furthermore, since the auxiliary predictions do not participate in the model inference phase, the FPS of the detectors will not drop after the introduction of the auxiliary losses.

\begin{figure}[t]
\centering
\includegraphics[width=0.85\textwidth]{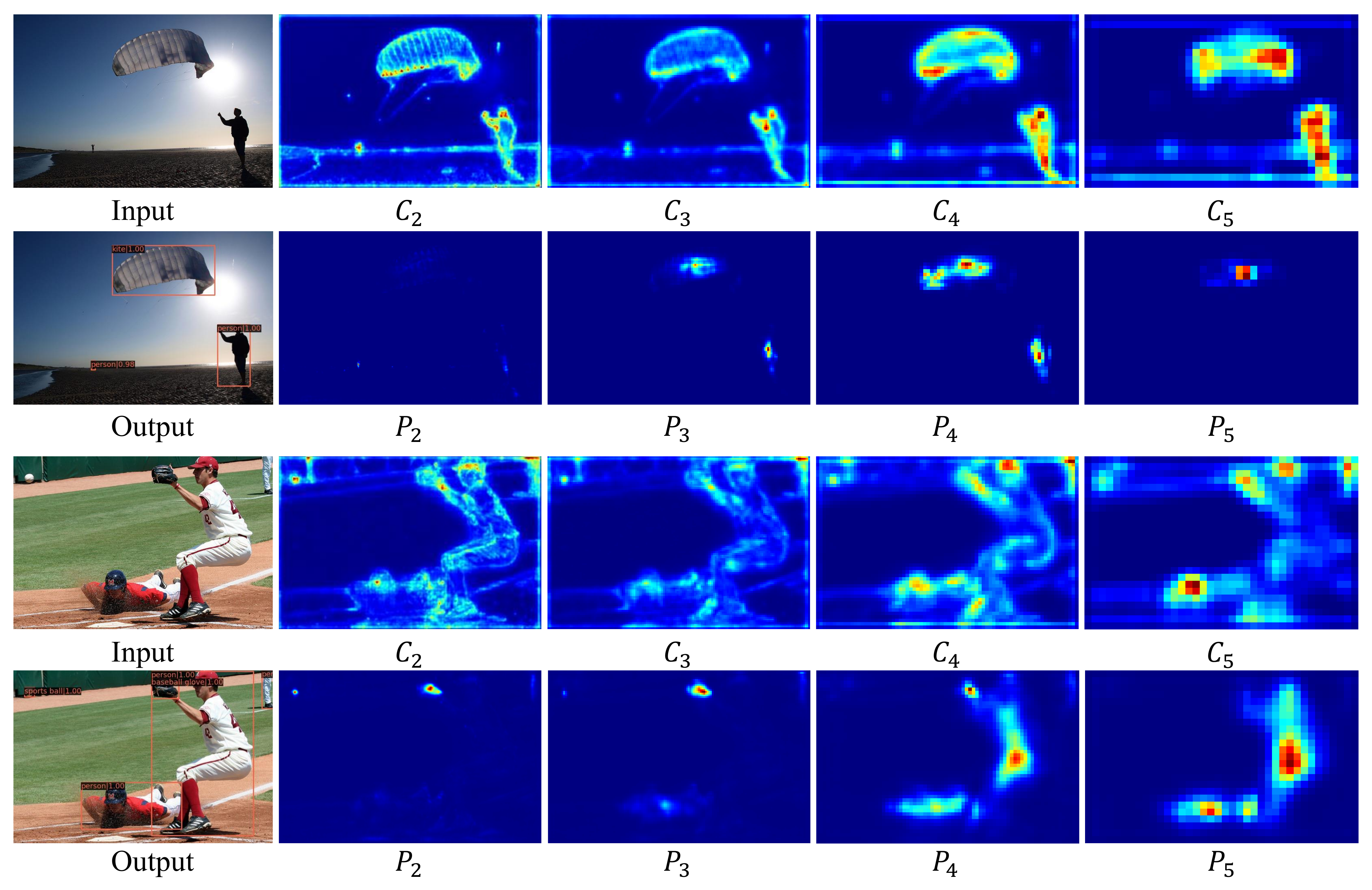}
\vspace{-0.2cm}
\caption{
   Visualization of the learned features. 
   The adopted model is Faster R-CNN with ResNet-101.
   The pictures are selected from MS COCO \emph{val-2017}.
}\label{fig.visfeatures}
\vspace{-0.6cm}
\end{figure}

\noindent \textbf{Feature Grouping.}
From the perspective of optimization, we have identified the unreasonable operation in the process of building traditional FPN.
Specifically, the top-down architecture will make the shallow layers of the backbone network fail to see the large-scale objects. 
To mitigate the adverse effects caused, we propose to leverage a feature grouping module to construct each feature pyramid level by selecting the feature maps from all backbone stages. 
Table \ref{table.ablationfg} demonstrates the empirical study.
We can observe that the FPN with feature grouping outperforms the traditional FPN by $1.7\%$ and $1.1\%$ in one-stage and two-stage detection framework, respectively.
And AP increases with the consistent rises in $\{\text{AP}_S,\text{AP}_M,\text{AP}_L\}$.
The result indicates that the feature grouping module has the ability to make each backbone level see all objects through amending the back-propagation paths between objective functions and the backbone network.

\noindent \textbf{Cascade Structure.}
To achieve more robust improvements by enhancing the space compactness of the homogeneous feature maps, we propose to introduce a cascade feature grouping structure.
The experimental results in Table \ref{table.ablationfg} demonstrate the effectiveness of this structure.
It is observed that the detection performance improves steadily as the number of cascade times increases in both one-stage and two-stage frameworks.
Moreover, the improvements of AP always benefit from the consistent improvements of $\{\text{AP}_S,\text{AP}_M,\text{AP}_L\}$.

\noindent \textbf{Visualization of Learned Features.}
In Figure~\ref{fig.visfeatures}, we visualize the feature maps outputted by the backbone levels (\emph{i.e.}, $\mathcal{C}$) and pyramid levels (\emph{i.e.}, $\mathcal{P}$).
It is observed that $\mathcal{C}$ contain the semantic information of the whole image, while $\mathcal{P}$ only carries the effective semantics used to detect the objects within the corresponding scale range,
indicating that the feature grouping module can well promise the space compactness of the homogeneous feature maps.

\begin{table*}[t]
\centering
\caption{
  The improvements on AP after integrating the cascade feature grouping module into various detection frameworks.
  The $1\times$, $3\times$ training schedules follow the settings explained in MMDetection~\cite{chen2019mmdetection}.
  FPN-CFG denotes for applying the cascade feature grouping module into FPN.
}\label{table.improvements}
\resizebox{.95\textwidth}{!}{
\begin{tabular}{c|c|c|cccccccc}
   \hline
   \hline
   Method                                    &Backbone                       &Schedule           &AP                                     &AP$_{50}$       &AP$_{75}$       &AP$_{S}$      &AP$_{M}$     &AP$_{L}$   \\
   \hline
   \emph{one-stage} &&&&&&&&\\
   RetinaNet \cite{lin2017focal}             &ResNet-101-FPN                 &$1\times$          &38.5                                   &57.6            &41.0            &21.7          &42.8        &50.4      \\
   RetinaNet \cite{lin2017focal}             &ResNet-101-FPN-CFG              &$1\times$          &41.2 (\textbf{{\color{teal}+2.7}})     &60.8            &43.8            &24.1          &45.2        &55.2      \\
   FreeAnchor \cite{zhang2021learning}       &ResNet-101-FPN                 &$1\times$          &40.3                                   &59.0            &43.1            &21.8          &44.0        &54.2      \\
   FreeAnchor \cite{zhang2021learning}       &ResNet-101-FPN-CFG              &$1\times$          &43.2 (\textbf{{\color{teal}+2.9}})     &62.0            &46.3            &24.4          &47.4        &57.6      \\
   ATSS \cite{zhang2020bridging}             &ResNet-101-FPN                 &$1\times$          &41.5                                   &59.9            &45.2            &24.2          &45.9        &53.3      \\
   ATSS \cite{zhang2020bridging}             &ResNet-101-FPN-CFG              &$1\times$          &43.8 (\textbf{{\color{teal}+2.3}})     &62.1            &47.3            &26.8          &48.0        &57.2      \\
   \hline
   \emph{two-stage} &&&&&&&&\\
   Faster R-CNN \cite{ren2015faster}         &ResNet-101-FPN                 &$1\times$          &39.4                                   &60.1            &43.1            &22.4          &43.7        &51.1      \\
   Faster R-CNN \cite{ren2015faster}         &ResNet-101-FPN-CFG              &$1\times$          &42.2 (\textbf{{\color{teal}+2.8}})     &63.0            &45.8            &25.5          &46.1        &55.8      \\
   Mask R-CNN \cite{he2017mask}              &ResNet-101-FPN                 &$1\times$          &40.0                                   &60.5            &44.0            &22.6          &44.0        &52.6      \\
   Mask R-CNN \cite{he2017mask}              &ResNet-101-FPN-CFG              &$1\times$          &43.3 (\textbf{{\color{teal}+3.3}})     &63.7            &47.6            &25.7          &47.1        &56.6      \\
   Cascade R-CNN \cite{cai2019cascade}       &ResNet-101-FPN                 &$1\times$          &42.0                                   &60.4            &45.7            &23.4          &45.8        &55.7      \\
   Cascade R-CNN \cite{cai2019cascade}       &ResNet-101-FPN-CFG              &$1\times$          &44.5 (\textbf{{\color{teal}+2.5}})     &63.1            &48.4            &26.1          &48.5        &57.8      \\
   Cascade Mask R-CNN \cite{cai2019cascade}  &ResNet-101-FPN                 &$1\times$          &42.9                                   &61.0            &46.6            &24.4          &46.5        &57.0      \\
   Cascade Mask R-CNN \cite{cai2019cascade}  &ResNet-101-FPN-CFG              &$1\times$          &45.4 (\textbf{{\color{teal}+2.5}})     &63.8            &49.4            &27.5          &49.3        &59.5      \\
   \hline
   \emph{anchor-free} &&&&&&&&\\
   FCOS \cite{tian2019fcos}                  &ResNet-50-FPN                  &$1\times$          &36.6                                   &56.0            &38.8            &21.0          &40.6        &47.0      \\
   FCOS \cite{tian2019fcos}                  &ResNet-50-FPN-CFG               &$1\times$          &39.6 (\textbf{{\color{teal}+3.0}})     &58.8            &42.3            &22.9          &43.4        &51.9      \\
   Sparse R-CNN \cite{sun2021sparse}         &ResNet-50-FPN                  &$1\times$          &37.9                                   &56.0            &40.5            &20.7          &40.0        &53.5      \\
   Sparse R-CNN \cite{sun2021sparse}         &ResNet-50-FPN-CFG               &$1\times$          &40.1 (\textbf{{\color{teal}+2.2}})     &58.7            &42.6            &22.2          &42.6        &55.6      \\
   FSAF \cite{zhu2019feature}                &ResNet-101-FPN                 &$1\times$          &39.3                                   &58.6            &42.1            &22.1          &43.4        &51.2      \\
   FSAF \cite{zhu2019feature}                &ResNet-101-FPN-CFG              &$1\times$          &42.2 (\textbf{{\color{teal}+2.9}})     &62.0            &44.8            &24.3          &45.9        &56.2      \\
   \hline
   \emph{transformer} &&&&&&&&\\
   Mask R-CNN \cite{liu2021swin}             &Swin-T-FPN                     &$1\times$          &42.7                                   &65.2            &46.8            &26.5          &45.9        &56.6      \\
   Mask R-CNN \cite{liu2021swin}             &Swin-T-FPN-CFG                  &$1\times$          &46.0 (\textbf{{\color{teal}+3.3}})     &67.0            &50.5            &28.8          &49.7        &59.1     \\
   \hline
   \emph{strong baseline} &&&&&&&&\\
   Cascade Mask R-CNN \cite{cai2019cascade}  &ResNeXt-101-64x4d-FPN	        &$3\times$           &46.6                                   &65.1            &50.6            &29.3          &50.5        &60.1      \\
   Cascade Mask R-CNN \cite{cai2019cascade}  &ResNeXt-101-64x4d-FPN-CFG  	  &$3\times$           &50.1 (\textbf{{\color{teal}+3.5}})     &68.6            &54.5            &32.7          &53.7        &64.3     \\
   \hline
   \hline
\end{tabular}}
\vspace{-0.7cm}
\end{table*}

\subsection{Performance with Various Detection Frameworks.}
To further prove the soundness of our principle and the robustness of our approach, we integrate the cascade feature grouping (CFG) structure into various detection frameworks.
Table~\ref{table.improvements} demonstrates the experimental results.
For one-stage detectors, our approach consistently improves the baseline frameworks by at least $2.3\%$ AP.
For two-stage detectors with pre-defined anchors and ResNet backbone, the baseline frameworks are increased by more than $2.5\%$ AP.
Recent academic attention has been geared toward anchor-free detectors and transformer-based backbone networks.
We have also made attempts to integrate the proposed structure into these frameworks.
It is observed that the cascade feature grouping structure brings more than $2.2\%$ AP improvements to the anchor-free detectors and the transformer-based detectors.
Moreover, we have also trained a strong baseline with multi-scale training, $3 \times$ schedule and ResNeXt-101-64x4d backbone. 
After integrating the cascade feature grouping module into the traditional FPN, the strong baseline is still improved by $3.5\%$ AP.
It is worth mentioning that the performance gains are all achieved by consistently boosting the AP of the objects within different scale ranges.
The results above together show the necessity and effectiveness that each level of the backbone network should own the ability to look at all objects.

\begin{table*}[t]
\centering
\caption{
  Experimental results on instance segmentation task.
  The models are trained on the MS COCO \emph{train-2017} split and evaluated on the MS COCO \emph{val-2017} set.
}\label{tab.improvements_seg}
\vspace{-0.2cm}
\resizebox{.95\textwidth}{!}{
\begin{tabular}{c|c|c|cccccccc}
   \hline
   \hline
   Method                                    &Backbone               &Schedule           &AP$^{seg}$                             &AP$^{seg}_{50}$       &AP$^{seg}_{75}$       &AP$^{seg}_{S}$      &AP$^{seg}_{M}$     &AP$^{seg}_{L}$   \\
   \hline
   Mask R-CNN                                &Swin-T-FPN             &$1\times$          &39.30                                  &62.20                 &42.20                 &20.50               &41.80              &57.80      \\
   Mask R-CNN                                &Swin-T-FPN-CFG        &$1\times$          &41.40 (\textbf{{\color{teal}+2.1}})    &64.50                 &44.60                 &21.90               &44.60              &58.80      \\
   \hline
   Mask R-CNN                                &ResNet-101-FPN         &$1\times$          &36.10                                  &57.50                 &38.60                 &18.80               &39.70              &49.50      \\
   Mask R-CNN                                &ResNet-101-FPN-CFG    &$1\times$          &38.70 (\textbf{{\color{teal}+2.6}})    &60.80                 &41.50                 &19.00               &42.00              &56.00      \\
   \hline
   Cascade Mask R-CNN                        &ResNet-101-FPN         &$1\times$          &37.30                                  &58.20                 &40.10                 &19.70               &40.60              &51.50      \\
   Cascade Mask R-CNN                        &ResNet-101-FPN-CFG    &$1\times$          &39.40 (\textbf{{\color{teal}+2.1}})    &61.30                 &42.60                 &19.70               &42.60              &57.10      \\
   \hline
   \hline
\end{tabular}}
\vspace{-0.2cm}
\end{table*}

\subsection{Instance Segmentation}
To verify the generalization ability of our approach, we also apply the cascade feature grouping module on a more challenging instance segmentation task, which requires the prediction of object instances and their per-pixel segmentation mask simultaneously.
As shown in Table~\ref{tab.improvements_seg}, our method improves AP$^{seg}$ of different detectors from $39.30\%$ to $41.40\%$, $36.10\%$ to $38.70\%$, and $37.30\%$ to $39.40\%$, respectively.
Moreover, all the improvements are built upon the consistent increases in $\{\text{AP}^{seg}_S,\text{AP}^{seg}_M,\text{AP}^{seg}_L\}$.

\section{Conclusions}
This work first identifies the nature of the success of FPN from the perspective of optimization.
Based on the principle, we succeed in illustrating the reason why the introduction of FPN will suppress the detection performance of large objects.
We further conclude that the key to address the inconsistent changes problem in $\{\text{AP}_S,\text{AP}_M,\text{AP}_L\}$ is to enable each backbone level to look at all objects.
Therefrom, we propose to design two strategies to achieve this goal.
One is to introduce the auxiliary losses so that the auxiliary signals containing the information of all objects can directly pass through the shallow layers of the backbone network.
The other is to integrate the cascade feature grouping structure into the existing FPN, which can also amend the back-propagation paths between the objective functions and the shallow layers of the backbone network.
Extensive experiments show the soundness of our principle and the effectiveness of our strategies.
Without bells and whistles, our method brings consistent performance improvements to 12 different detection frameworks.




\clearpage
%
%
\bibliographystyle{splncs04}
\bibliography{YSLAO}
\end{document}